%% file: templateArxiv.tex
\documentclass{article}

\newcommand*\samethanks[1][\value{footnote}]{\footnotemark[#1]}
\usepackage{PRIMEarxiv}

\usepackage[utf8]{inputenc} 
\usepackage[T1]{fontenc}    
\usepackage{hyperref}       
\usepackage{url}            
\usepackage{booktabs}       
\usepackage{amsfonts}       
\usepackage{nicefrac}       
\usepackage{microtype}      
\usepackage{lipsum}		
\usepackage{graphicx}
\usepackage{doi}
\usepackage{bbm}
\usepackage{bm}
\usepackage{enumitem}
\usepackage{wrapfig}
\usepackage{xspace}
\usepackage{natbib}
\setcitestyle{numbers,square}
\usepackage{booktabs}
\usepackage{mathtools}
\usepackage{multicol}
\usepackage{algorithm}
\usepackage{algorithmicx}
\usepackage{algpseudocode}
\usepackage{color-edits}
\usepackage{amsfonts,amsthm,amsmath}
\usepackage{rotating}

\usepackage{color-edits}

\addauthor[Sixie]{sixie}{blue} 
\addauthor[zonglin]{zongl}{red}
\addauthor[yang]{yangl}{red}

\newcommand{\squishlist}{
\begin{list}{{{\small{$\bullet$}}}}
{\setlength{\itemsep}{3pt}      \setlength{\parsep}{1pt}
\setlength{\topsep}{1pt}       \setlength{\partopsep}{0pt}
\setlength{\leftmargin}{1em} \setlength{\labelwidth}{1em}
\setlength{\labelsep}{0.5em} } }
\newcommand{\squishend}{  \end{list}  }

\title{Adversarial Machine Unlearning
}

\author{
  Zonglin Di\thanks{For equal contribution} \\
  UC Santa Cruz \\
  \texttt{zdi@ucsc.edu} \\
   \And
  Sixie Yu\samethanks \\
  Stellar Cyber, Inc.\\
  \texttt{bit.yusixie@gmail.com} \\
   \And
  Yevgeniy Vorobeychik \\
  Washington University in St. Louis \\
  \texttt{yvorobeychik@wustl.edu} \\
   \And
  Yang Liu \\
  UC Santa Cruz \\
  \texttt{yangliu@ucsc.edu} \\
}

\DeclareMathOperator{\argmax}{arg\,max}

\DeclareMathSymbol{\R}{\mathord}{AMSb}{"52}
\DeclarePairedDelimiter\norm{\lVert}{\rVert}

\providecommand{\SET}[1]{\ensuremath{\{ #1 \}}\xspace}          
\providecommand{\Set}[2]{\ensuremath{\SET{#1 \mid #2}}\xspace}  
\providecommand{\Abs}[1]{\ensuremath{\left| #1 \right|}\xspace} 

\providecommand{\data}{\ensuremath{ D }\xspace} 
\providecommand{\rdata}{\ensuremath{ D_r }\xspace} 
\providecommand{\trdata}{\ensuremath{ D_{tr} }\xspace} 
\providecommand{\valdata}{\ensuremath{ D_{val} }\xspace} 
\providecommand{\tedata}{\ensuremath{ D_{te} }\xspace} 
\providecommand{\fdata}{\ensuremath{ D_{f} }\xspace} 
\providecommand{\MIAdata}{\ensuremath{ \tilde{D} }\xspace} 
\providecommand{\X}{\ensuremath{ \mathcal{X} }\xspace} 
\providecommand{\Y}{\ensuremath{ \mathcal{Y} }\xspace} 
\providecommand{\MIAnum}{\ensuremath{ q }\xspace} 
\providecommand{\xf}{\ensuremath{ x^f }\xspace} 
\providecommand{\xte}{\ensuremath{ x^{te} }\xspace} 
\providecommand{\floss}{\ensuremath{ \mathcal{L}_f }\xspace} 
\providecommand{\teloss}{\ensuremath{ \mathcal{L}_{te} }\xspace} 

\providecommand{\costd}{\ensuremath{ C_u }\xspace} 
\providecommand{\utila}{\ensuremath{ U_a }\xspace} 
\providecommand{\tradeoff}{\ensuremath{ \alpha }\xspace} 

\providecommand{\unlearnF}{\ensuremath{ \theta_u }\xspace} 
\providecommand{\origF}{\ensuremath{ \theta_o }\xspace} 
\providecommand{\retainF}{\ensuremath{ \theta_r }\xspace} 
\providecommand{\outputF}{\ensuremath{ s^f }\xspace} 
\providecommand{\outputTE}{\ensuremath{ s^{te} }\xspace} 
\providecommand{\attackerF}{\ensuremath{ \theta_a }\xspace} 
\providecommand{\unlearnerF}{\ensuremath{ \theta_u }\xspace} 
\providecommand{\HypoA}{\ensuremath{ \mathcal{H}_a }\xspace} 
\providecommand{\HypoD}{\ensuremath{ \mathcal{H}_u }\xspace} 
\providecommand{\BestR}[1]{\ensuremath{ \mathcal{B}_{#1} }\xspace} 
\providecommand{\Output}[1]{\ensuremath{ S_{#1} }\xspace} 
\providecommand{\Model}{\ensuremath{ \theta }\xspace}  
\providecommand{\ProxyModel}{\ensuremath{ \tilde{\theta} }\xspace} 

\providecommand{\ML}{\ensuremath{ \mathcal{A} }\xspace} 
\providecommand{\MU}{\ensuremath{ \mathcal{U} }\xspace} 
\providecommand{\KKT}{\ensuremath{ f }\xspace} 

\providecommand{\Lobj}{\ensuremath{ V }\xspace} 
\providecommand{\Loss}{\ensuremath{ L }\xspace} 
\providecommand{\Veri}{\ensuremath{ M }\xspace} 
\providecommand{\Acc}{\ensuremath{Acc}\xspace}

\makeatletter

\newcounter{ALG@prevlno}
\makeatother

\newcommand{\itemTitle}[1]{\item \textbf{#1:}}

\theoremstyle{remark}

\begin{document}
\maketitle

\begin{abstract}
This paper focuses on the challenge of machine unlearning, aiming to remove
the influence of specific training data on machine learning models. 
Traditionally, the development of unlearning algorithms runs parallel with that of membership inference attacks (MIA), a type of privacy threat to determine whether a data instance was used for training. 
However, the two strands are intimately connected: one can view machine unlearning through the lens of MIA success with respect to removed data.
Recognizing this connection, we propose a game-theoretic framework that integrates MIAs into the design of unlearning algorithms.
Specifically, we model the unlearning problem as a Stackelberg game in which an unlearner strives to unlearn specific training data from a model, while an auditor employs MIAs to detect the traces of the ostensibly removed data.
Adopting this adversarial perspective allows the utilization of new attack advancements,  facilitating the design of unlearning algorithms.
Our framework stands out in two ways. 
First, it takes an adversarial approach and proactively incorporates the attacks into the design of unlearning algorithms. 
Secondly, it uses implicit differentiation to obtain the gradients that limit the attacker's success, thus benefiting the process of unlearning.
We present empirical results to demonstrate the effectiveness of the proposed approach for machine unlearning.
\end{abstract}


\section{Introduction}
\label{sec:intro}
\input{sections/intro}

\section{Related Work}
\label{sec:related_work}
\input{sections/related}

\section{Preliminaries}
\label{sec:prelim}
\input{sections/prelim}
\section{The Game Model}
\label{sec:model}
\input{sections/model}

\section{\texttt{SG-Unlearn}: Stackelberg Game Unlearn}
\label{sec:algo}
\input{sections/algo}

\section{Experiments}
\label{sec:exp}
\input{sections/exp}

\section{Discussion}
\label{sec:disc}
\input{sections/disc}

\bibliographystyle{unsrt}  
\bibliography{references}

\newpage
\appendix

\input{sections/appendix}

\end{document}

%% file: sections/intro.tex
The enactment of the General Data Protection Regulation (GDPR) by the EU has elevated the importance of deleting user data from machine learning models to a critical level. 
This process is distinctly more intricate compared to removing data from conventional databases. 
Erasing the data's imprint from a machine learning model necessitates an approach to negate the data's influence on the model comprehensively while maintaining the utility and accuracy of the model.

Beyond this, establishing the true extent to which data influence has been erased from the model poses a significant challenge~\citep{song2021systematic}. 
Numerous methods and metrics have been advanced to validate the thoroughness of data removal, each with varying degrees of reliability and efficacy~\citep{guo2020certified,thudi22unlearn}. 
We propose a novel adversarial perspective on unlearning that we argue is a more robust framework for effective machine unlearning.
In this approach, the focus shifts to simulating possible attacks aimed at inferring whether the data that should have been forgotten nevertheless maintains some influence on the model.
If, within this adversarial framework, an attacker fails to distinguish whether a data point was part of the training set or merely a typical instance of unseen data, we can conclude that the influence of the data point on the data has been successfully unlearned. 

We leverage advancements from the burgeoning domain of Membership Inference Attacks (MIA) to simulate an adversary~\citep{MIA:shokri2017membership}, therein framing a Stackelberg Game (SG) between an unlearner, tasked with orchestrating the unlearning process, and an auditor deploying MIA to deduce the membership of data in the model's training set.
The key idea is for the unlearner to adjust the model being unlearned by utilizing gradient feedbacks from the auditor's optimization problem, moving the model in a direction that limits the effectiveness of the attack, thus achieving the goal of unlearning.
Specifically, we formulate the MIA as a utility-maximizing problem, where the utility measures the remaining influence of a data point in the unlearned model. 
The unlearner's loss function is defined as a combination of the degradation of model performance and the auditor's utility. 
We harness the development from implicit differentiation and design a gradient-based algorithm to solve the game, allowing for seamless integration into existing end-to-end pipelines~\citep{Gould2016diff,amos2017optnet,Agrawal2019diffcvx, wang2023zeroth}.

The contributions of the present paper are summarized below
\squishlist
    \item We propose to evaluate the effectiveness of an unlearning algorithm from an adversarial perspective, inspiring us to develop a game theory framework that enables the use of advanced MIAs for enhancing the unlearning process. 
    \item Additionally, we design a gradient-based solution method to solve the game by leveraging implicit differentiation, making it amenable to end-to-end pipelines.
    \item Finally, we support the efficacy of the game and the solution method with extensive results. 
\squishend


%% file: sections/related.tex
The first related thread is machine unlearning, which focuses on removing the influence of a subset of data (referred to as the forget set) from a machine learning model.
The unlearning approaches are divided into two classes. The first one is exact unlearning, which involves retraining the model on data excluding the forget set.
The second one is approximate unlearning. 
The ideas behind approximate unlearning are twofold.
The first is to track the influence of each training data on the updates to a model's weights, allowing for a rollback during unlearning~\citep{bourtoule2021machine,graves21amne,chen2022graph}.
The second is using a loss function to capture the objectives of unlearning (e.g., removing the influence of the forget set while maintaining model utility) and  modifying the model weights to minimize the loss function~\citep{guo2020certified,golatkar20fforget,izzo21deletion,Warnecke23mu,chundawat23teacher,jia23mu}.
The method proposed in this paper aligns with the second idea.
Specifically, we design a loss function that simulates an auditor who uses MIAs to evaluate the effectiveness of unlearning.
By differentiating through the auditor's optimization problem, we compute the gradients that reduce the auditor's utility, thus increasing the effectiveness of unlearning. 
Besides algorithmic developments, \cite{Jagielski2023measure} proposes a measure to quantify the forgetting during training; \cite{thudi22unlearn} takes a formal analysis of the definition of approximation unlearning and propose methods to verify exact unlearning.
Due to space constraint, it is not feasible to provide a comprehensive review of all related studies. We refer the readers to the survey article by~\cite{nguyen22unlearn} for a more exhaustive discussion.

The second related line is membership inference attacks (MIA).
\cite{MIA:shokri2017membership} introduced MIAs,  showing the privacy risks of machine learning models.
Subsequently, different attack methods are proposed~\citep{chen2021jeop,MIA:carlini2022membership,ye2022mia,bertran23quantile}.
On the other hand, \cite{MIA:carlini2022membership} shows that existing criteria  to evaluate MIAs are limited in capturing real-world scenarios and propose more practical evaluation metrics. 
In addition, comprehensive evaluation frameworks and tools are developed~\citep{sasi2020meter,song2021systematic}. 
Finally, \cite{nasr2018machine} proposes a defense mechanism to counter MIAs from an adversarial perspective.
Our method shares conceptual similarities with this work, but there are several key differences. Our primary focus is on machine unlearning problems, while their focus is on defending against MIAs. This means that our framework needs to support multiple types of MIAs to provide a comprehensive evaluation of unlearning, including both neural network (NN)-based and non-NN-based attacks. However, their framework only supports NN-based attacks. Furthermore, NN-based attacks are generally not suitable for our runtime requirements; indeed, if unlearning takes longer than retraining, we would opt for retraining instead.

%% file: sections/prelim.tex
\noindent \textbf{Machine Unlearning.}
Let $\data=\Set{(x_i, y_i)}{x_i \in \X, y_i \in \Y}$ be a labeled dataset, where \X (resp. \Y) denote the feature (resp. label) space. 
The training, validation, and test sets are $\trdata$, $\valdata$, and $\tedata$, respectively.
A machine learning (ML)  algorithm is denoted by \ML, mapping from the joint space of features and labels $\X \times \Y$ to a hypothesis class.
We refer to the model trained on the entire training set as the original model, i.e., $\origF = \ML(\trdata)$.

Let $\fdata = \SET{(x_{j}^f, y_{j}^f)}_{j=1}^{\MIAnum} \subseteq \trdata$ represent a forget set. 
The goal of machine unlearning is to remove the influence of \fdata from the original model, resulting in an unlearned model \unlearnF (i.e., $\unlearnF = \MU(\origF)$) where $\MU$ represents a machine unlearning algorithm.
The unlearning algorithm may have access to other inputs (e.g., the validation set \valdata) depending on the problem settings.
Let \rdata be the retain set, the subset of the training data excluding the forget set, i.e.,  $\rdata = \trdata \setminus \fdata$.
The gold standard of machine unlearning is $\retainF = \ML(\rdata)$, a model trained on the retain set, excluding the influence of \fdata. 
We denote \retainF as the gold standard when comparing  machine unlearning algorithms.
Retraining is expensive, especially for deep neural networks.
This motivates the development of efficient machine unlearning algorithms that satisfy the following conditions: 1) the influence of \fdata is removed from the unlearned model, 2) the performance of the unlearned model is comparable to the performance of the original model, and 3) the computational costs (e.g., running time) are lower compared to those incurred during retraining.

\noindent \textbf{Membership Inference Attacks.}
A membership inference attack (MIA) aims to determine whether a data instance was used to train an ML model~\citep{MIA:shokri2017membership}. 
An instance that was in the training set is called a member, while one that was not in the training set is called a non-member. 
Formally, given a target model \Model, an attacker infers the membership of an instance $(x, y)$ based on the model's outputs (i.e., $\Output{\Model}(x)$) and the label.
The attacker does not have access to either the training data or the model parameters of the target model.
Instead, he gathers proxy training and test sets and learns a model \ProxyModel to mimic the behavior of the target model.
Using the outputs of \ProxyModel on its own training and test data, the attacker acquires a labeled (member v.s. non-member) dataset, and then uses the labeled dataset to train a binary classifier for determining the membership of an instance.

We adapt  the idea of MIA to determine whether the influence of the forget set still exists in an unlearned model \unlearnF.
Define an auditing set $\MIAdata_{\unlearnF} = \SET{(\outputF_j, 1), (\outputTE_j, 0)}_{j=1}^{\MIAnum}$, where $\outputF_j$ (resp. $\outputTE_j$)  represents the outputs of the forget (resp. test) instances from the unlearned model, that is,  $\outputF_j = \Output{\unlearnF}(\xf_j)$ (resp. $\outputTE_j = \Output{\unlearnF}(\xte_j)$).

Here, the test instances serve as an empirical distribution for the unseen data. 
The outputs can be scalars, such as the instance-wise cross-entropy losses.
The outputs can also be the vectors of probabilities across  the classes~\citep{MIA:shokri2017membership,MIA:carlini2022membership}.
The labels ``1" and ``0" indicate members and non-members, respectively.
The MIA reduces to a binary classification task on $\MIAdata_{\unlearnF}$, aiming to differentiate the forget instances from the test ones based on the outputs. 

%% file: sections/model.tex
We model the machine unlearning problem as a Stackelberg game (SG) between an unlearner who deploys models as services, and an auditor who launches MIAs against the  models.
The key idea is to assess the effectiveness of an unlearning algorithm by measuring whether the auditor will succeed. 
In particular, the unlearning is considered effective when the auditor is unable to differentiate between the forget set from the test set based on their outputs from the unlearned model.
The SG is played in a sequential manner: the unlearner first deploys an unlearned model, and then the auditor launches an MIA  in response.
Importantly, the advantage of first-mover endows the unlearner with the power to make a decision knowing that the auditor will play a best response (i.e., launching a strong attack).
We now formally define the models for both players.

\subsection{The Auditor's Model}\label{sec:attacker_model}
We begin by defining the auditor's model.
Suppose the unlearner has deployed an unlearned model $\unlearnF$. 
Following standard setup~\citep{MIA:shokri2017membership, song2021systematic}, 
we assume that the auditor has black-box access to the model, allowing him to query the model, e.g., submitting data to the model and collecting the outputs. 
The auditor's goal is to determine whether the influence of the forget set still exists in the model based on the outputs. 
To achieve this, the auditor constructs an auditing set $\MIAdata_{\unlearnerF}$, consisting of the model's outputs when passing the forget and test instances through the unlearning model \unlearnerF (see Section~\ref{sec:prelim} for details about the auditing set).
The auditor  assesses the distinctiveness of the two sets with a binary classifier trained on the auditing set through cross validation.

Let \utila be the auditor's utility function, quantifying the distinctiveness of the forget and test instances.
Intuitively, a large \utila indicates that the outputs of the forget instances are highly differentiable from the outputs of the test instances, strong evidence that the influence of the forget set still exists in the unlearned model.
We formulate the auditor's model as the following optimization problem
\vspace{-0.1in}
\begin{equation}
\label{eq:attacker_cost}
\begin{aligned}
    & \utila(\attackerF, \unlearnerF)  =\Veri(\MIAdata^{val}_{\unlearnerF}; \attackerF) ~~~~ \text{where } \attackerF \in \BestR{\unlearnerF} = \argmax_{\attackerF^\prime \in \HypoA} \Veri(\MIAdata^{tr}_{\unlearnerF}; \attackerF^\prime).
\end{aligned}
\end{equation}

The auditing set $\MIAdata_{\unlearnerF}$ is divided into the training $\MIAdata^{tr}_{\unlearnerF}$ and the validation $\MIAdata^{val}_{\unlearnerF}$ sets.
The constraint encodes the process of learning a binary classifier.
The set $\BestR{\unlearnerF}$ are the auditor's best-responses to the unlearner's decision \unlearnerF, that is, a specific MIA that maximally differentiates the forget and test instances. 
The function \Veri is an evaluation metric for the binary classifier on a dataset. 
The definition of \Veri is flexible.
One can use the accuracy to quantify the average performance of the classifier, where true positives are weighted equally with true negatives~\citep{MIA:shokri2017membership,song2021systematic}.
Alternatively, an average measure may not capture real privacy threats.
Instead, ROC curve or true positive rates at specified false positive rates 
can be used for evaluation~\cite{MIA:carlini2022membership}.

The auditor's model exhibits a high degree of generality, unifying several advanced MIAs in the literature; this includes neural network-based attacks proposed by~\cite{nasr2018machine}, quantile regression-based attacks from~\cite{bertram2019five}, and prediction confidence-based attacks by~\cite{song2021systematic}, etc.
Under the formulation of \eqref{eq:attacker_cost}, the mentioned attacks differ in 1) the hypothesis class \HypoA of the binary classifier  and 2) the objective function \Veri. 
Notice the dependence of the auditor's best-response  on \unlearnerF (i.e., $\BestR{\unlearnerF}$) arising from the unlearner's first-mover advantage. 
The unlearner utilizes this dependence to select an unlearned model that limits the auditor's discriminative power, which we discuss next.

\subsection{The Unlearner's Model}\label{sec:defender_model}
Next, we define the unlearner's model.
Let \costd represent the unlearner's cost function, which encompasses two main objectives for unlearning.
The first objective is to maintain the utility of the model, ensuring that the unlearned model performs comparably (e.g., in terms of predictive power) to the original model.
To achieve this objective, we minimize a loss function $\Loss(\rdata; \unlearnerF)$ computed on the retain set \rdata, following the principles of empirical risk minimization.
All regularization terms are included in the loss function to simplify notation.
The second objective focuses on eliminating the influence of the forget set from the model being unlearned.
We approach this objective adversarially by considering the auditor's utility \Veri.
In essence, a smaller value of the auditor's utility indicates that the forget set is harder to be distinguished from the test set, providing strong evidence that the unlearning process is effective.

Formally, the unlearner's optimization problem is to minimize the cost function below
    \begin{equation}\label{eq:defender_cost}
        \costd\left( \unlearnerF, \attackerF \right) = \Loss(\rdata; \unlearnerF) + \tradeoff \cdot \Veri(\MIAdata^{val}_{\unlearnerF}; \attackerF).
    \end{equation}
The parameter $\tradeoff \in \R^+$ balances the loss \Loss and the auditor's utility \Veri.
Depending on the specific setting, the cost function \costd can be extended to incorporate additional objectives for unlearning. For instance, one can specify that the unlearned model should perform poorly on the forget set~\citep{graves21amne}; this can be achieved by minimizing  an evaluation metric (e.g., likelihood) on the forget set.
Also, several sparsity-promoting  techniques have been shown helpful for unlearning~\citep{jia23mu}; one way to achieve this is by adding an $\ell_1$ regularization to the cost function.

\subsection{The Stackelberg Game}\label{sec:sg}
Now, with the unlearner and the auditor models in place, we  formally define the Stackelberg Game (SG).
The SG is to solve the following bi-level optimization problem~\citep{colson2007overview}
\begin{equation}\label{eq:adv_unlearn}
    \begin{aligned}
        & \min_{\unlearnerF\in \HypoD} & &  \underbrace{\Loss(\rdata; \unlearnerF) + \tradeoff \cdot \Veri(\MIAdata^{val}_{\unlearnerF}; \attackerF)}_{\text{Unlearner}}    ~~~~ s.t. \quad   \underbrace{\attackerF \in \BestR{\unlearnerF}}_{\text{Auditor}}.
    \end{aligned}
\end{equation}
The hierarchical structure encodes the sequential order of the play, with the upper level corresponding to the unlearner's optimization problem  and  the lower level  capturing the auditor's best-responses.
During the unlearning, the unlearner needs to proactively consider the auditor's responses. 
This requires selecting an unlearning model where the influence of the forget set is erased, or from the auditor's perspective, the forget instances are indistinguishable from the test ones.

The key assumption of the SG is that if the forget set cannot be distinguished from the test set---in terms of the effectiveness of an MIA---its influence is deemed eliminated from the unlearned model.
We justify this assumption from three angles. 
First, one common way to measure forgetfulness is by assessing the accuracy of the unlearned model on the forget set~\citep{graves21amne,chundawat23teacher,baumhauer2022machine}. 
This approach is grounded on the observation that machine learning models exhibit distinct performance between training data and unseen data.
However, it is important to note that accuracy on the forget set does not necessarily correlate with forgetfulness, as there are inherently difficult (or easy) instances that result in low (or high) accuracy regardless of whether they were part of the training set~\citep{MIA:carlini2022membership}.
Secondly, MIAs have been used to study training data forgetting~\citep{Jagielski2023measure}, demonstrating its utility in detecting residual traces of a dataset. 
Finally, from an adversarial perspective, if a sophisticated attack like an MIA cannot differentiate the forget set from the test set, it is reasonable to expect that the influence of the forget set has been removed.

We solve the SG using gradient-based methods, allowing for easy integration into end-to-end training pipelines. 
Specifically, we use  Implicit Function Theorem to differentiate through the auditor's optimization problem \eqref{eq:attacker_cost}, obtaining the gradient of the auditor's utility with respect to (w.r.t) the unlearning model's weights, i.e.,  $\partial \Veri / \partial \unlearnerF$. 
As a result, the SG  becomes a differentiable layer, compatible with the standard forward-backward computation. 
The solution methods will be detailed in the next section.

%% file: sections/algo.tex

In this section, we describe the algorithm for solving the SG.
In general, it is NP-hard to find an optimal solution for the unlearner~\citep{conitzer06sg}. 
Instead, we focus on gradient-based algorithms to find an approximate solution, i.e., a model parameter \unlearnerF exhibiting good unlearning performance. 
The main technical challenge is computing the gradient of the auditor's utility w.r.t. the unlearning model's weights (i.e., $\partial \Veri / \partial \unlearnerF$), which requires differentiation through the auditor's optimization problem. 
While the differentiation can be bypassed in some special cases, e.g., when the unlearner's hypothesis class is of linear regressions~\citep{tong2018adv},  this is rarely applicable in the current setting given our primary focus on unlearning deep neural networks. 

Our solution leverages both the Implicit Function Theorem (IFT)~\citep{dontchev2009implicit} and tools from Differentiable Optimization (DO) to compute the gradients~\citep{Gould2016diff,amos2017optnet,Agrawal2019diffcvx}, thereby rendering the SG a differentiable layer seamlessly integrable into existing end-to-end pipelines.

We start by expanding the gradient of \costd w.r.t. \unlearnerF using the chain rule
\begin{equation}
\label{eq:costd_grad}
    \frac{\partial \costd}{\partial \unlearnerF} = \frac{\partial \Loss(\rdata; \unlearnerF)}{\partial \unlearnerF} + \frac{\partial \Veri(\MIAdata^{val}_{\unlearnerF}; \attackerF)}{\partial \attackerF} \cdot \frac{\partial \attackerF }{\partial \MIAdata^{tr}_{\unlearnerF}} \cdot \frac{\partial \MIAdata^{tr}_{\unlearnerF}}{\partial \unlearnerF}.
    \end{equation}
The first term on the right-hand side can be easily computed using an automatic differentiation tool like PyTorch~\citep{paszke2017automatic}. In essence, the computation involves passing \rdata through the unlearning model (i.e., \unlearnerF) in the forward pass, computing the loss \Loss, and getting the gradients in the backward pass.
The second term on the right is an expansion of $\partial \Veri / \partial \unlearnerF$ using the chain rule; for clarity we omit the arguments of the functions.
The gradient $\partial \Veri / \partial \attackerF$ is obtained by performing a standard forward-backward pass.
Some evaluation metrics for binary classification, such as the 0-1 loss, AUC, recall, etc., are non-differentiable. Therefore, we adhere to standard practices by employing a differentiable proxy for \Veri, such as utilizing the logistic loss as a substitute for the 0-1 loss.

\noindent{\bf Leveraging Implicit Function Theorem } Computing the gradient $\partial \attackerF / \partial \MIAdata^{tr}_{\unlearnerF}$ requires differentiation through the attacker's optimization problem.
The main challenge is the absence of an explicit function that maps $\MIAdata^{tr}_{\unlearnerF}$ to $\attackerF$. 
However, under certain regularity assumptions, one can derive an implicit mapping between $\MIAdata^{tr}_{\unlearnerF}$ and $\attackerF$ based on the optimality conditions of the auditor's optimization problem~\citep{Gould2016diff}.
A concrete example is when the optimization problem is convex, such as learning a Support Vector Machine (SVM) \footnote{This includes several state-of-the-art MIAs~\citep{bertran23quantile,song2021systematic}.}.
In this case, the KKT conditions are necessary and sufficient conditions for the optimality, and it 
connects $\attackerF$ with $\MIAdata^{tr}_{\unlearnerF}$ through a system of linear equations, i.e., 
    \begin{equation}\label{eq:kkt}
        \KKT(\MIAdata^{tr}_{\unlearnerF}, \attackerF) = 0,
    \end{equation}
where \KKT encapsulates the stationarity conditions, the primal and dual feasibility conditions, and the complementary slackness conditions~\citep{boyd2004convex}. 
For illustration purposes, a concrete example of the KKT conditions \KKT for linear SVM is provided in Appendix~\ref{app:implicit_diff}.
We apply IFT to the system of linear equations, resulting in
\begin{equation}
\label{eq:implicit}
    \frac{\partial \attackerF}{\partial \MIAdata^{tr}_{\unlearnerF}} = -\left( \frac{\partial\KKT(\MIAdata^{tr}_{\unlearnerF}, \attackerF) }{\partial \MIAdata^{tr}_{\unlearnerF}}\right)^{-1} \frac{\partial \KKT(\MIAdata^{tr}_{\unlearnerF}, \attackerF)}{\partial \attackerF}.
\end{equation}
For further insights into differentiating through an optimization problem using the implicit function theorem, we recommend referring to the lectures by~\cite{gould2023lecture}.


\noindent{\bf Leveraging Differentiable Optimization } In practice, we capitalize on tools from Differentiable Optimization (DO) to compute the gradients.
Intuitively, we can consider DO as software that implements IFT, as shown in equation \eqref{eq:implicit}, for a given optimization problem.
What we need to do is describing the auditor's optimization problem using a specialized modeling language, e.g., \texttt{cvxpy}~\citep{diamond2016cvxpy}.
We then use DO to transform this description into a differentiable layer. 
Subsequently, this differentiable layer is positioned atop the model undergoing unlearning, thereby establishing a computational pathway from \unlearnerF to \attackerF. 
The pseudocode for this process is provided in Algorithm~\ref{algo:sg_algo}. 
This algorithm has a time complexity of $O(n^3)$, where $n$ denotes the size of the attacker's optimization problem (i.e., the number of variables and/or constraints). This cubic dependence stems from the matrix inversion in~\eqref{eq:implicit}.


\vspace{-0.1in}
\begin{algorithm}
\caption{\texttt{SG-Unlearn}}\label{algo:sg_algo}
    \begin{algorithmic}[1]
        \State Input:  $\rdata, \fdata, \tedata$ and the original model $\origF$
        \State Initialize: $i=0, \unlearnerF^0 = \origF$, a scheduler $\eta^i$
        \While{$i <$ epoch}
        \State Compute $\Loss(\rdata; \unlearnerF^i)$ on the retain set in a forward pass
        \State Update $\textcolor{red}{\unlearnerF^{i^{\prime} } } \leftarrow \unlearnerF^i - \eta^i  \cdot \frac{\partial \Loss(\rdata; \unlearnerF^i)}{\partial \unlearnerF^i}$
        
        \State Construct the auditing set $\MIAdata_{\textcolor{red}{\unlearnerF^{i^{\prime}}} }$ from $\fdata$ and $\tedata$
        \State Describe the auditor's optimization problem \eqref{eq:attacker_cost} with \texttt{cvxpy}
        \State Convert the description to a differentiable layer \texttt{AuditorLayer}
        \State Plug \texttt{AuditorLayer} into the computational graph
        \State Get the auditor's best response $\attackerF^i \gets \texttt{AuditorLayer}(\MIAdata_{\textcolor{red}{\unlearnerF^{i^{\prime}}} })$
        \State Compute $\Veri\left( \MIAdata^{val}_{ \textcolor{red}{\unlearnerF^{i^{\prime}}}  }; \attackerF^i \right)$
        \State Update $\unlearnerF^{i+1} \gets \textcolor{red}{\unlearnerF^{i^{\prime}}}  - \eta^i \cdot \frac{\partial \Veri \left(\MIAdata^{val}_{ \textcolor{red}{\unlearnerF^{i^{\prime}}}  }; \attackerF^i\right) }{\partial \unlearnerF^i}$ 
        \State $i \gets i + 1$
        \EndWhile
        \State Return: $\unlearnerF^i$
    \end{algorithmic}
\end{algorithm}

%% file: sections/exp.tex
\subsection{Experiment Setup}

We run experiments on CIFAR-10, CIFAR-100, and SVHN, three well-known image classification datasets~\citep{krizhevsky2009learning,netzer2011reading}.
For all experiments we use the ResNet-18 architecture~\citep{he2016deep}.
We consider both \textit{random forgetting} and \textit{class-wise forgetting} 
In random forgetting, instances are sampled uniformly at random from all classes. 
In contrast, class-wise forgetting entails selecting all instances from a specific class.
For CIFAR-10 and CIFAR-100, the forget set consists of 10\% of the entire training set, while the ratio is 5\%  for SVHN.
In all experiments, the attacker's optimization problem is formulated as a binary classification task, wherein a linear support vector machine (SVM) is employed to distinguish between forget and test instances.

The baseline methods we use to compare with SG include Retrain\footnote{Retraining the unlearning model on $\trdata \setminus \fdata$ from scratch.}, Fine-Tune (FT) \citep{warnecke2021machine, golatkar2020eternal}, Gradient Ascent (GA) \citep{graves2021amnesiac, thudi2022unrolling}, Influence Unlearning (IU) \citep{izzo2021approximate, koh2017understanding}, $\ell_1$-sparse \citep{jia23mu}, Random Label (RL) \citep{hayase2020selective}, Boundary Expansion (BE), Boundary Shrink (BS) \citep{chen2023boundary} and SCRUB \citep{kurmanji2024towards}
Further details on the baseline methods are provided in Section~\ref{sec:baseline_methods} of the Appendix.
For all methods, we use the SGD optimizer with a weight decay of 5e-4 and a momentum of 0.9.
Other hyper-parameters are selected through the validation set.
Specifically, we create a new auditing set. For each unlearning method, we select the hyper-parameters that maximize the difference between the validation accuracy and the MIA accuracy on this new auditing set.
The hyperparameters are listed in Table~\ref{tab:hyperparameters} in the Appendix.

\subsection{Evaluation Metrics}
We evaluate SG and the baseline methods using metrics previously adopted in related studies~\citep{bourtoule2021machine, Jagielski2023measure, jia23mu, chundawat23teacher}. 
\emph{It is important to note that the test accuracy is evaluated on a subset of the test data that is separate from the one used for solving SG.}
\textbf{Retain accuracy (${\Acc}_{r}$) and test accuracy (${\Acc}_{te}$)} are used to quantify model utility~\citep{jia23mu}.
\textbf{MIA accuracy, AUC and F1 score} are the metrics to quantify the effectiveness of unlearning, which are estimated on the auditing set with 10-fold cross \cite{MIA:carlini2022membership}.
A good unlearning algorithm should be close to random guessing (0.5) in terms of the MIA metrics.
\textbf{Forget accuracy (${\Acc}_f$)} measures the accuracy of the unlearned model on the forget set. 
An effective unlearning algorithm should produce an unlearned model where ${\Acc}_f$ is close to the ${\Acc}_{te}$.
Indeed, if the unlearned model did not see $\fdata$, its performance on this set should be consistent with that on $\tedata$.
\textbf{The absolute difference between the forget and test accuracy (|${\Acc}_{f}-{\Acc}_{te}$|)} measures whether the unlearned model performs consistently on $\fdata$ and $\tedata$.
To gather additional statistical evidence regarding the effectiveness of unlearning, we collect the cross-entropy losses of the forget and test instances from the unlearned model into the empirical distributions \floss and \teloss, respectively.
Next, we run a \textbf{Kolmogorov-Smirnov statistics (KS Stat.)} test to determine if the distributions can be differentiated from each other.
The KS statistic quantifies the differences between \floss and \teloss, where the p-value indicates whether the difference is significant~\citep{massey1951kolmogorov}.
In addition to the KS statistics, we provide the \textbf{Wasserstein distance (W. Dist.)} between the empirical distributions of \floss and \teloss.
This complements the KS statistics and evaluates the unlearning performance in terms of the similarity between the losses.

\subsection{Results}
The experimental results of random forgetting and class-wise forgetting are presented in Section~\ref{sec:random_forgetting} and \ref{sec:class_forgetting}, respectively.
We consider retrain as the gold standard for evaluating  unlearning algorithms: the closer to the metrics of retrain the more effective the algorithm. 
We highlight the closest metrics to retrain in bold. 

\subsubsection{Random Forgetting}
\label{sec:random_forgetting}

We present the results of CIFAR-10 and CIFAR-100 in Table \ref{tab:random_forgetting_main}. 
The results for SVHN datasets are provided in \ref{app:svhn_result} in the Appendix.
SG achieves the best performance for most of the metrics, demonstrating its effectiveness in unlearning. 
Specifically, the KS statistic of SG is consistently lower than those of the other baselines, exhibiting an order of magnitude difference in the statistics for CIFAR-10 compared to most baselines.
Intuitively, ML models behave differently on training data compared to unseen data, and this difference is usually reflected in the corresponding losses~\citep{MIA:carlini2022membership}. 
The small KS statistic of SG implies that the forget and test instances exhibit greater similarity in terms of the model's behavior, although there is still a discernible difference between the losses.
Another metric for measuring the similarity is the Wasserstein distance (W. Dist.). 
The baseline RL achieves the lowest distance, although the difference with SG is not statistically significant.
A visualization of the cross-entropy losses for the forget and test instances from one of the experiments is provided in Figure~\ref{fig:loss_dist} in the Appendix.

Another observation from the table is that there is a trade-off between model performance, measured by test accuracy, and the effectiveness of unlearning, measured by MIA accuracy.
Specifically, SG is more effective at unlearning the forget instances, as indicated by the highlighted MIA metrics.
However, this effectiveness comes at a cost to the test accuracy on CIFAR-10 and CIFAR-100, although the degradation is not significant.
We run a large array of experiments with varying  \tradeoff from $\{0.05, 0.1, 0.25, 0.5, 1, 2, 5\}$ (see \eqref{eq:defender_cost}) to explore the extent to which the trade-off can be reduced.
The results are presented in Figure~\ref{fig:alpha_main}.
Unfortunately, we do not see a consistent trend that makes SG closer to retrain across all the metrics. 

\subsubsection{Class-wise Forgetting}
\label{sec:class_forgetting}
We present CIFAR-10 here as the benchmark of class-wise forgetting. 
{For other datasets, please refer to Section \ref{app:class_wise} in the Appendix.}
In contrast to random forgetting, where the forget set is sampled uniformly at random across all classes, class-wise forgetting involves removing the influence of all instances belonging to a specific class.
The experimental results are shown in Table \ref{tab:class_wise_main}.
The metrics closest to retrain are highlighted. 
Notably, all methods perform poorly on MIA-related metrics. In other words, the auditor can readily distinguish between forget instances and test instances. This outcome is expected: unlike random forgetting, class-wise forgetting produces significantly different distributions between the forgotten instances (e.g., fruits) and the test instances (e.g., non-fruits). Consequently, the outputs of the unlearning model are expected to be easily distinguishable. 
Additionally, no single method consistently outperforms the others.

\vspace{-0.1in}
\begin{table}[ht]
    \centering
    \caption{Experimental results (Mean$_{\text{std}}$) on CIFAR-10 and CIFAR-100 for random forgetting. The highlighted metrics are the closest to those of retraining, which is considered as the best performance compared with the other baselines. 
    }
    \scalebox{0.65}{
    \begin{tabular}{c|cccc|cccccc}
    \toprule
     CIFAR-10 & ${\Acc}_{r}$  & ${\Acc}_{te}$ & ${\Acc}_{f}$ & $\Abs{{\Acc}_{f} - {\Acc}_{te}}$ & MIA acc. & MIA AUC & MIA F1 & KS Stat. & W. Dist. & RTE (min., $\downarrow$) \\
    \midrule
    Retrain & 0.9996$_{0.0001}$ & 0.9291$_{0.0022}$ & 0.9230$_{0.0043}$ & 0.0061 & 0.5069$_{0.0073}$ & 0.5083$_{0.0099}$ & 0.5094$_{0.0187}$ & 0.0255$_{0.0080}$ & 0.0307$_{0.0115}$ & 14.92\\
    \midrule
    FT & 0.9886$_{0.0055}$ & 0.9114$_{0.0050}$ & 0.9851$_{0.0056}$ & 0.0737 & 0.5405$_{0.0031}$ & 0.5457$_{0.0070}$ & 0.6293$_{0.0127}$ & 0.0933$_{0.0050}$ & 0.3158$_{0.0142}$ & 0.45\\
    GA & \textbf{0.9996}$_{0.0001}$ & 0.9304$_{0.0007}$ & 0.9995$_{0.0003}$ & 0.0691 & 0.5504$_{0.0066}$ & 0.5611$_{0.0071}$ & 0.6625$_{0.0106}$ & 0.1403$_{0.0037}$ & 0.2782$_{0.0026}$ & 0.18\\
    IU & 0.9723$_{0.0255}$ & 0.8966$_{0.0242}$ & 0.9722$_{0.0243}$ & 0.0756 & 0.5398$_{0.0055}$ & 0.5548$_{0.0076}$ & 0.6193$_{0.0204}$ & 0.1050$_{0.0192}$ & 0.4083$_{0.0554}$ & 0.02\\
    $\ell_1$-sparse & 0.9970$_{0.0007}$ & 0.9234$_{0.0014}$ & 0.9938$_{0.0016}$ & 0.0704 & 0.5501$_{0.0049}$ & 0.5694$_{0.0087}$ & 0.6405$_{0.6259}$ & 0.1018$_{0.0034}$ & 0.2704$_{0.0074}$ & 0.96\\
    RL & 0.9988$_{0.0001}$ & 0.9217$_{0.0008}$ & 0.9810$_{0.0025}$ & 0.0593 & 0.5217$_{0.0087}$ & 0.5297$_{0.0133}$ & \textbf{0.5935}$_{0.0140}$ & 0.0986$_{0.0136}$ & \textbf{0.1520}$_{0.0058}$ & 0.84\\
    BE & \textbf{0.9996}$_{0.0001}$ & 0.9304$_{0.0007}$ & 0.9996$_{0.0003}$ & 0.0692 & 0.5541$_{0.0049}$ & 0.5639$_{0.0058}$ & 0.6629$_{0.0082}$ & 0.1412$_{0.0030}$ & 0.2783$_{0.0020}$ & 0.27\\
    BS & 0.9995$_{0.0001}$ & 0.9307$_{0.0008}$ & 0.9995$_{0.0004}$ & 0.0688 & 0.5588$_{0.0072}$ & 0.5779$_{0.0097}$ & 0.6590$_{0.0156}$ & 0.1466$_{0.0032}$ & 0.3072$_{0.0026}$ & 0.46\\
    SCRUB & 0.9971$_{0.0018}$ & \textbf{0.9251}$_{0.0018}$ & 0.9959$_{0.0022}$ & 0.0708 & 0.5533$_{0.0059}$ & 0.5679$_{0.0073}$ & 0.6337$_{0.0149}$ & 0.1038$_{0.0071}$ & 0.2485$_{0.0154}$ & 1.30\\
    \midrule
    SG & 0.9948$_{0.0029}$ & 0.8940$_{0.0048}$ & \textbf{0.9351}$_{0.0070}$ & \textbf{0.0411} & \textbf{0.5202}$_{0.0054}$ & \textbf{0.5134}$_{0.0084}$ & 0.6480$_{0.0043}$ & \textbf{0.0482}$_{0.0082}$ & 0.1555$_{0.0194}$ & 1.47 \\
    \midrule
    \midrule
    CIFAR-100 & ${\Acc}_{r}$  & ${\Acc}_{te}$ & ${\Acc}_{f}$ & $\Abs{{\Acc}_{f} - {\Acc}_{te}}$ & MIA acc. & MIA AUC & MIA F1 & KS Stat. & W. Dist. & RTE (min., $\downarrow$) \\
    \midrule
    Retrain & 0.9996$_{0.0001}$ & 0.7035$_{0.0025}$ & 0.6925$_{0.0039}$ & 0.0110 & 0.5184$_{0.0057}$ & 0.5281$_{0.0053}$ & 0.5104$_{0.0081}$ & 0.0203$_{0.0045}$ & 0.0567$_{0.0200}$ & 13.08\\
    \midrule
    FT & 0.9991$_{0.0001}$ & 0.7117$_{0.0021}$ & 0.9984$_{0.0006}$ & 0.2867 & 0.6630$_{0.0075}$ & 0.7300$_{0.0102}$ & 0.6878$_{0.0107}$ & 0.4566$_{0.0083}$ & 1.2583$_{0.0166}$ & 0.39\\
    GA & \textbf{0.9996}$_{0.0001}$ & 0.7158$_{0.0008}$ & 0.9996$_{0.0002}$ & 0.2838 & 0.6977$_{0.0060}$ & 0.7601$_{0.0065}$ & 0.7207$_{0.0088}$ & 0.4915$_{0.0030}$ & 1.2219$_{0.0038}$ & 0.20 \\
    IU & 0.9971$_{0.0029}$ & \textbf{0.7026}$_{0.0080}$ & 0.9959$_{0.0034}$ & 0.2933 & 0.6660$_{0.0089}$ & 0.7305$_{0.0134}$ & 0.6950$_{0.0124}$ & 0.4583$_{0.0168}$ & 1.2612$_{0.0366}$ & 0.21\\
    $\ell_1$-sparse & 0.9958$_{0.0013}$ & 0.7095$_{0.0025}$ & 0.9890$_{0.0028}$ & 0.2785 & 0.6738$_{0.0081}$ & 0.7392$_{0.0088}$ & 0.6952$_{0.0073}$ & 0.3717$_{0.0113}$ & 1.1157$_{0.0079}$ & 0.84\\
    RL & 0.9965$_{0.0054}$ & 0.6665$_{0.0031}$ & 0.8483$_{0.0447}$ & 0.1818 & 0.5808$_{0.0426}$ & 0.6152$_{0.0527}$ & 0.6080$_{0.0466}$ & 0.2323$_{0.0731}$ & 0.8067$_{0.1705}$ & 0.73\\
    BE & 0.9995$_{0.0001}$ & 0.7173$_{0.0014}$ & 0.9996$_{0.0002}$ & 0.2823 & 0.6977$_{0.0037}$ & 0.7661$_{0.0066}$ & 0.7248$_{0.0065}$ & 0.4940$_{0.0031}$ & 1.2175$_{0.0119}$ & 0.23\\
    BS & 0.9995$_{0.0001}$ & 0.7160$_{0.0013}$ & 0.9996$_{0.0002}$ & 0.2836 & 0.6987$_{0.0052}$ & 0.7651$_{0.0072}$ & 0.7239$_{0.0054}$ & 0.4963$_{0.0033}$ & 1.2382$_{0.0206}$ & 0.39\\
    SCRUB & 0.9993$_{0.0001}$ & 0.7097$_{0.0019}$ & 0.9991$_{0.0004}$ & 0.2894 & 0.7015$_{0.0057}$ & 0.7747$_{0.0060}$ & 0.7299$_{0.0056}$ & 0.4717$_{0.0052}$ & 1.2280$_{0.0128}$ & 1.14\\
    \midrule
    SG & 0.8993$_{0.0105}$ & 0.6378$_{0.0066}$ & \textbf{0.7239}$_{0.0093}$ & \textbf{0.0861} & \textbf{0.5412}$_{0.0070}$ & \textbf{0.5320}$_{0.0076}$ & \textbf{0.6061}$_{0.0056}$ & \textbf{0.0988}$_{0.0061}$ & \textbf{0.5316}$_{0.0295}$ & 3.07 \\
    \bottomrule
    \end{tabular}
    }
    \label{tab:random_forgetting_main}
\end{table}

\begin{table}[htb!]
    \centering
    \caption{Experimental results (Mean$_{\text{std}}$) on CIFAR-10 for class-wise forgetting. The highlighted metrics are the closest to those of retraining, which is considered as the best performance compared with the other baselines.}
    \scalebox{0.65}{
    \begin{tabular}{c|cccc|cccccc}
    \toprule
     CIFAR-10 & ${\Acc}_{r}$  & ${\Acc}_{te}$ & ${\Acc}_{f}$ & $\Abs{{\Acc}_{f} - {\Acc}_{te}}$ & MIA acc. & MIA AUC & MIA F1 & KS Stat. & W. Dist. & RTE (min., $\downarrow$) \\
    \midrule
    Retrain & 0.9996$_{0.0001}$ & 0.9333$_{0.0009}$ & 0.0000$_{0.0000}$ & 0.9333 & 0.9935$_{0.0006}$ & 0.9983$_{0.0004}$ & 0.9936$_{0.0007}$ & 0.9803$_{0.0002}$ & 9.5601$_{0.0911}$ & 13.96 \\
    \midrule
    FT & 0.9958$_{0.0022}$ & 0.9226$_{0.0030}$ & 0.6043$_{0.0450}$ & 0.3183 & 0.9915$_{0.0011}$ & \textbf{0.9985$_{0.0002}$} & 0.9915$_{0.0011}$ & 0.7975$_{0.0133}$ & 0.9324$_{0.1847}$ & 1.16 \\
    GA & 0.8478$_{0.0046}$ & 0.7942$_{0.0055}$ & 0.0007$_{0.0002}$ & 0.7935 & \textbf{0.9944$_{0.0011}$} & 0.9996$_{0.0002}$ & \textbf{0.9938$_{0.0015}$} & 0.9269$_{0.0087}$ & 15.2941$_{0.1656}$ & 0.84\\
    IU & 0.9339$_{0.0161}$ & 0.8644$_{0.0141}$ & 0.0619$_{0.0149}$ & 0.8025 & 0.9972$_{0.0009}$ & 0.9996$_{0.0001}$ & 0.9972$_{0.0007}$ & 0.8151$_{0.0195}$ & \textbf{8.2574$_{0.6484}$} & 0.31 \\
    $\ell_1$-sparse & 0.9972$_{0.0005}$ & 0.9285$_{0.0014}$ & 0.0914$_{0.0310}$ & 0.8371 & 0.9910$_{0.0014}$ & 0.9989$_{0.0001}$ & 0.9910$_{0.0014}$ & 0.9208$_{0.0078}$ & 2.5552$_{0.1738}$ & 1.84 \\
    RL & \textbf{0.9996$_{0.0000}$} & \textbf{0.9330$_{0.0008}$} & 0.0001$_{0.0001}$ & \textbf{0.9329} & 0.9916$_{0.0013}$ & 0.9990$_{0.0005}$ & 0.9916$_{0.0013}$ & 0.9695$_{0.0025}$ & 6.3989$_{0.0789}$ & 1.97 \\
    BE & 0.9710$_{0.0012}$ & 0.8984$_{0.0023}$ & 0.2477$_{0.0022}$ & 0.6507 & 0.9964$_{0.0005}$ & 0.9990$_{0.0004}$ & 0.9964$_{0.0005}$ & 0.7306$_{0.0047}$ & 4.8984$_{0.0432}$ & 0.32 \\
    BS & 0.9691$_{0.0031}$ & 0.8969$_{0.0031}$ & 0.2504$_{0.0105}$ & 0.6465 & 0.9965$_{0.0001}$ & 0.9988$_{0.0005}$ & 0.9965$_{0.0002}$ & 0.7196$_{0.0072}$ & 5.0155$_{0.0922}$ & 0.66 \\
    SCRUB & 1.0000$_{0.0000}$ & 0.9269$_{0.0021}$ & \textbf{0.0000$_{0.0000}$} & 0.9269 & 1.0000$_{0.0000}$ & 1.0000$_{0.0000}$ & 1.0000$_{0.0000}$ & 0.9999$_{0.0001}$ & 70.9934$_{2.9441}$ & 3.47\\
    \midrule
    SG & 0.9667$_{0.0054}$ & 0.9056$_{0.0055}$ & \textbf{0.0000$_{0.0000}$} & 0.9056 & 0.9814$_{0.0026}$ & 0.9902$_{0.0025}$ & 0.9818$_{0.0025}$ & \textbf{0.9696$_{0.0032}$} & 5.2754$_{0.1882}$ & 0.84\\
    \bottomrule
    \end{tabular}
    }
    \label{tab:class_wise_main}
\end{table}

\subsubsection{The Effect of the Attacker Model}
\begin{figure}[ht] 
    \centering 
    \includegraphics[width=0.95\textwidth]{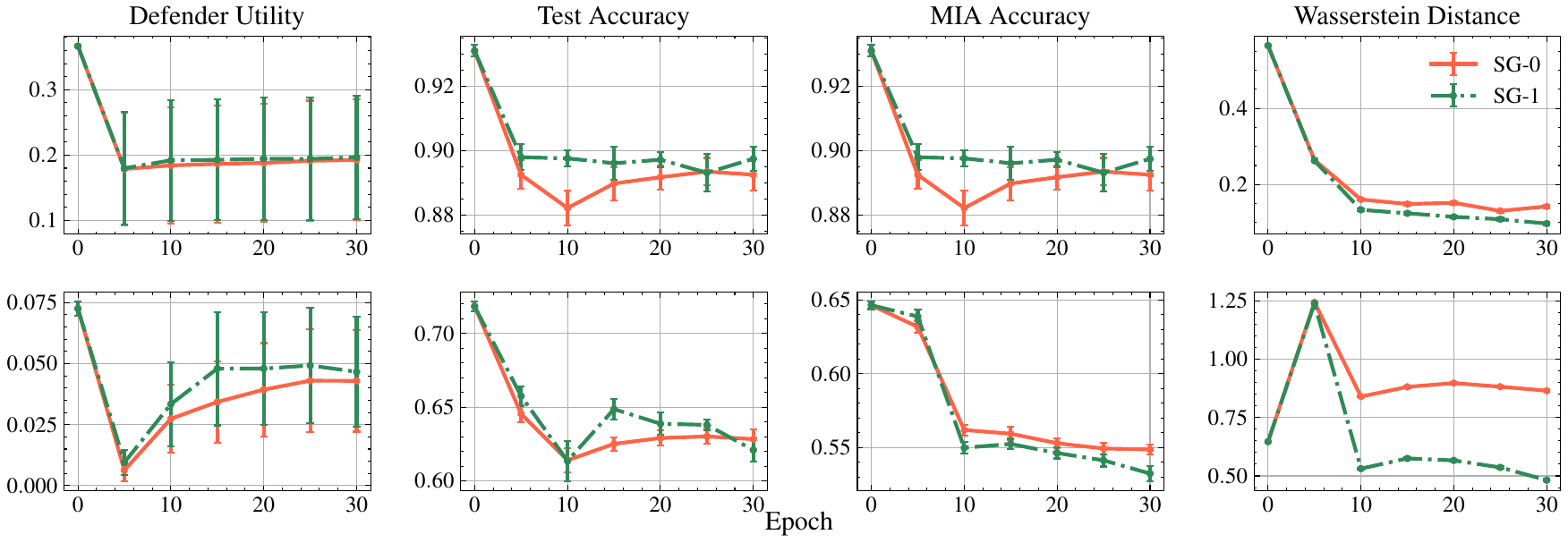}
    \caption{An ablation study to understand the impact of adversarial modeling on the process of unlearning; $\tradeoff=1$ and $\tradeoff=0$ corresponds to the cases with and without adversarial modeling, respectively.  
    The results are the averages over 10 experiments with different seeds, and 95\% confidence intervals are displayed. 
    \textbf{From the left to the right}: 1) the defender's utility, evaluated as the test accuracy ${\Acc}_{te}$ minus the MIA accuracy; 2) test accuracy; 3) MIA accuracy; 4) Wasserstein distance between the cross-entropy losses of the forget and test instances. 
    \textbf{Top row}: CIFAR-10; \textbf{Bottom row}: CIFAR-100.
    \textbf{Epoch 0: Original model.}
    }
    \label{fig:ablation}
\end{figure}

Finally, we conduct a comparative study to understand the impact of adversarial modeling on the unlearning process, controlled by the parameter $\tradeoff$ as defined in \eqref{eq:defender_cost}.
We show the results for random forgetting and defer the results for class-wise forgetting to the appendix. 
In Figure~\ref{fig:ablation}, we compare two cases where \tradeoff is set to either 1 or 0, denoted by SG-1 and SG-0 respectively.
The comparison is done across four metrics:  1) the test accuracy ; 2) the MIA accuracy; 3) the defender's utility, evaluated as the test accuracy minus the MIA accuracy, which provides a combined scalar value that measures both the performance of the unlearned model and the effectiveness of unlearning; 4) the Wasserstein distance between the empirical distributions of \floss and \teloss.
We show the averages over 10 experiments with different seeds, and 95\% confidence intervals are displayed. 
The first observation is that the adversarial term (i.e., $\tradeoff \cdot \Veri(\MIAdata^{val}_{\unlearnF}; \attackerF)$) acts as a regularizer, improving the generalizability of the unlearned model.
This observation is supported by comparing the test accuracy of SG-1 and SG-0 on CIFAR-10 (top middle).
Similar findings have been reported in~\cite{nasr2018machine}.
Another observation is that adversarial modeling limits the attacker's ability to differentiate between forget instances and test instances; this is demonstrated by the MIA accuracy on CIFAR-100.
The right-most column displays the Wasserstein distances between \floss and \teloss.
It is evident that the two losses are closer as a result of adversarial modeling, especially for CIFAR-100 dataset. 
Additionally, the distances progressively decrease throughout the epochs, confirming the effectiveness of the gradient-based method.

\begin{figure}[ht]
    \centering
    \includegraphics[width=0.95\textwidth]{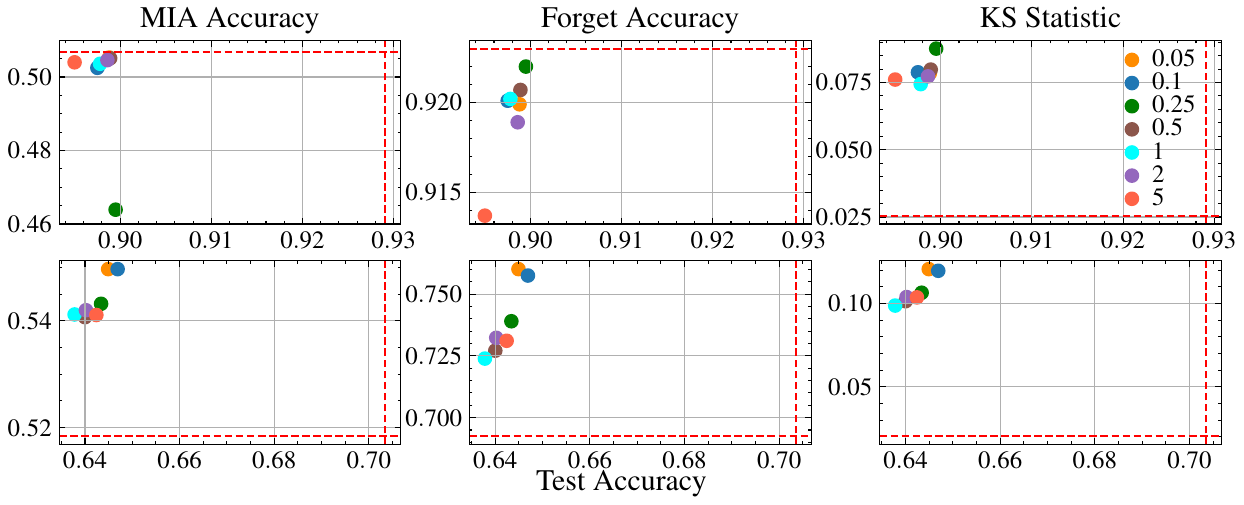}
    \caption{Experiments with different values of the trade-off parameter \tradeoff.
    We consider 7 values \SET{0.05, 0.1, 0.25, 0.5, 1, 2, 5}.
    Each dot represents a batch of 5 random experiments with the same \tradeoff.
    The coordinates of a dot are the corresponding metrics averaged over the 5 runs. 
    \textbf{Top row}: CIFAR-10; \textbf{Bottom row}: CIFAR-100.}
    \label{fig:alpha_main}
\end{figure}

In addition to the existence of attacker, we also investigate the strength of the attacker by changing $\tradeoff$. We select the $\tradeoff$ in large range of $\{0.05, 0.1, 0.25, 0.5, 1, 2, 5\}$. In Figure \ref{fig:alpha_main}, we compare the performance regarding the test accuracy $\Acc_{te}$. The cross of the red dash line is the performance of the retrain model. We can find that SG is robust to the attacker strength.

\vspace{-0.1in}

%% file: sections/disc.tex
In this paper, we design an adversarial framework for addressing the problem of unlearning a set data from a machine learning model.
Our approach focuses on evaluating the effectiveness of unlearning from an adversarial perspective, leveraging membership inference attacks (MIAs) to detect any residual traces of the data within the model.
The framework allows for a proactive design of the unlearning algorithm, synthesizing two lines of research---machine unlearning and MIAs---that have heretofore progressed in parallel.
By using implicit differentiation techniques, we develop a gradient-based algorithm for solving the game, making the framework easily integrable into existing end-to-end learning pipelines.
We present empirical results to support the efficacy of the framework and the algorithm.
We believe our work can make a progress in trustworthy ML.


\textbf{Limitation}
One future direction is to enhance the algorithm's efficiency. As shown in \eqref{eq:implicit}, the gradient-based algorithm requires a matrix inversion, which exhibits an $O(n^3)$ dependence on the size of the auditor's optimization problem. Therefore, developing a more efficient method to differentiate through the auditor's optimization could significantly accelerate the algorithm.
Another direction is to experiment with different combinations of the unlearning algorithm and the MIA within the SG framework. Currently, the unlearner employs Fine-Tune as the unlearning algorithm, while the auditor uses an SVM-based MIA. Exploring the performance of other combinations, such as Random Label with a neural network-based MIA, would be worthwhile.

%% file: sections/appendix.tex
\section{Appendix}

\subsection{Notation Table}
\begin{table}[htb!]
    \centering
    \caption{A summary of the notations used in the paper}
    \scalebox{1}{
    \begin{tabular}{c|l}
    \toprule
    Notation & Meaning \\
    \midrule
        $\mathcal{D}=\{(x_i, y_i)\}$ & A dataset \\
        $(x_i, y_i)$ & One data point where $x_i$ is the feature while $y_i$ is the label \\
        $\mathcal{X}, \mathcal{Y}$ & The feature space and the label space \\
        $D_f, D_{te}, D_{val}, D_{tr}, D_{r}$ & The forgetting, testing, validation, training, and retain set \\
        $(x_j^f, y_j^f)$ & One data point from the forget set $D_f$ \\
        $\mathcal{A}$ & A machine learning algorithm \\
        $\mathcal{U}$ & A machine unlearning algorithm \\
        $\theta_o$ & The original model, i.e., $\mathcal{A}(D_{tr})$ \\
        $\theta_u$ & The unlearned model, i.e., $\mathcal{U}(\theta_o)$ \\
        $\theta_r$ & The retrained model, i.e., $\mathcal{A}(D_{r})$ \\
        $\MIAdata_{\unlearnF}$ & The auditing dataset for membership inference attack \\
        $s_{j}^{f}$ & The output of a forget instance in the auditing dataset $\tilde{D}_{\theta_u}$ from $\theta_u$ \\
        $s_{j}^{te}$ & The output of a testing instance in the auditing dataset $\tilde{D}_{\theta_u}$ from $\theta_u$ \\
        $\MIAdata_{\unlearnF}^{tr}, \MIAdata_{\unlearnF}^{val}$ & The training and validation split of $\MIAdata_{\unlearnF}$. \\
        $\costd$ & The unlearner's cost function \\
        $\BestR{\unlearnerF}$ & The auditor's best response given an unlearning model \unlearnerF \\
        $\utila$ & The utility function of the auditor \\
        $\HypoA$ & The hypothesis class of the auditor \\
        $\HypoD$ & The hypothesis class of the unlearner \\
        $\alpha$ & The trade-off factor as defined in the unlearner's cost function \eqref{eq:defender_cost} \\
    \bottomrule
    \end{tabular}
    }
    \label{tab:notation}
\end{table}

\subsection{Random Forgetting}

\subsubsection{SVHN Dataset}\label{app:svhn_result}
The results of SVHN dataset on random forgetting are given in Table \ref{tab:svhn_random}.

\begin{table}[ht]
    \centering
    \caption{Experimental results (Mean$_{\text{std}}$) on SVHN for random forgetting. The highlighted metrics are the closest to those of retraining, which is considered as the best performance compared with the other baselines.}
    \scalebox{0.65}{
    \begin{tabular}{c|cccc|cccccc}
    \toprule
     SVHN & ${\Acc}_{r}$  & ${\Acc}_{te}$ & ${\Acc}_{f}$ & $\Abs{{\Acc}_{f} - {\Acc}_{te}}$ & MIA acc. & MIA AUC & MIA F1 & KS Stat. & W. Dist. & RTE (min., $\downarrow$) \\
     \midrule
     Retrain & 0.9959$_{0.0002}$ & 0.9610$_{0.0010}$ & 0.9534$_{0.0024}$ & 0.0076 & 0.5248$_{0.0058}$ & 0.5422$_{0.0075}$ & 0.5149$_{0.0157}$ & 0.0306$_{0.0117}$ & 0.0686$_{0.0145}$ & 20.46 \\
     \midrule
     FT & 0.9991$_{0.0001}$ & 0.7117$_{0.0021}$ & 0.9876$_{0.0070}$ & 0.2867 & 0.5372$_{0.0123}$ & 0.5592$_{0.0121}$ & 0.5523$_{0.0170}$ & 0.0613$_{0.0342}$ & 0.1743$_{0.0091}$ & 1.55\\
     GA & 0.9954$_{0.0001}$ & 0.9641$_{0.0002}$ & 0.9949$_{0.0006}$ & 0.0308 & 0.5191$_{0.0072}$ & 0.5411$_{0.0051}$ & 0.5500$_{0.0178}$ & 0.0867$_{0.0065}$ & 0.1473$_{0.0026}$ & 0.97\\
     IU & 0.9076$_{0.0707}$ & 0.8817$_{0.0658}$ & 0.9050$_{0.0713}$ & 0.0233 & 0.5373$_{0.0116}$ & 0.5580$_{0.0097}$ & 0.5469$_{0.0187}$ & 0.0473$_{0.0207}$ & 0.1407$_{0.0882}$ & 0.41\\
     $\ell_1$-sparse & 0.9378$_{0.0615}$ & 0.9191$_{0.0540}$ & 0.9298$_{0.0620}$ & 0.0107 & 0.5457$_{0.0220}$ & 0.5665$_{0.0229}$ & 0.5347$_{0.0324}$ & \textbf{0.0396}$_{0.0112}$ & 0.1158$_{0.0954}$ & 1.86\\
     RL & 0.9949$_{0.0002}$ & \textbf{0.9609}$_{0.0006}$ & 0.9797$_{0.0018}$ & 0.0188 & 0.5211$_{0.0106}$ & 0.5411$_{0.0147}$ & \textbf{0.5144$_{0.0225}$}& 0.1079$_{0.0175}$ & \textbf{0.0642}$_{0.0060}$ & 2.65 \\
     BE & 0.9955$_{0.0001}$ & 0.9633$_{0.0002}$ & 0.9955$_{0.0006}$ & 0.0322 & 0.5209$_{0.0090}$ & 0.5441$_{0.0064}$ & 0.5553$_{0.0175}$ & 0.1016$_{0.0062}$ & 0.1528$_{0.0019}$ & 0.46\\
     BS & \textbf{0.9956}$_{0.0002}$ & 0.9641$_{0.0001}$ & 0.9952$_{0.0008}$ & 0.0311 & 0.5322$_{0.0060}$ & 0.5509$_{0.0034}$ & 0.5594$_{0.0176}$ & 0.0994$_{0.0074}$ & 0.1404$_{0.0033}$ & 0.81 \\
     SCRUB & 0.9832$_{0.0010}$ & 0.9559$_{0.0014}$ & 0.9809$_{0.0020}$ & 0.0250 & \textbf{0.5273$_{0.0031}$} & \textbf{0.5431$_{0.0103}$} & 0.5296$_{0.0196}$ & 0.0492$_{0.0139}$ & 0.1032$_{0.0141}$ & 1.85\\
     \midrule
     SG & 0.9686$_{0.0017}$ & 0.9576$_{0.0033}$ & \textbf{0.9560}$_{0.0027}$ & \textbf{0.0016} & 0.5012$_{0.0052}$ & 0.5089$_{0.0272}$ & 0.3292$_{0.1798}$ & 0.0594$_{0.0233}$ & 0.0185$_{0.0041}$ & 3.16\\
     \bottomrule
    \end{tabular}
    }
    \label{tab:svhn_random}
\end{table}

\begin{figure}
    \centering
    \caption{An ablation study to understand the impact of adversarial modeling on the process of unlearning; $\tradeoff=1$ and $\tradeoff=0$ corresponds to the cases with and without adversarial modeling, respectively.  
    The results are the averages over 10 experiments with different seeds, and 95\% confidence intervals are displayed. 
    \textbf{From the left to the right}: 1) the defender's utility, evaluated as the test accuracy ${\Acc}_{te}$ minus the MIA accuracy; 2) test accuracy; 3) MIA accuracy; 4) Wasserstein distance between the cross-entropy losses of the forget and test instances.}
    \includegraphics[width=\textwidth]{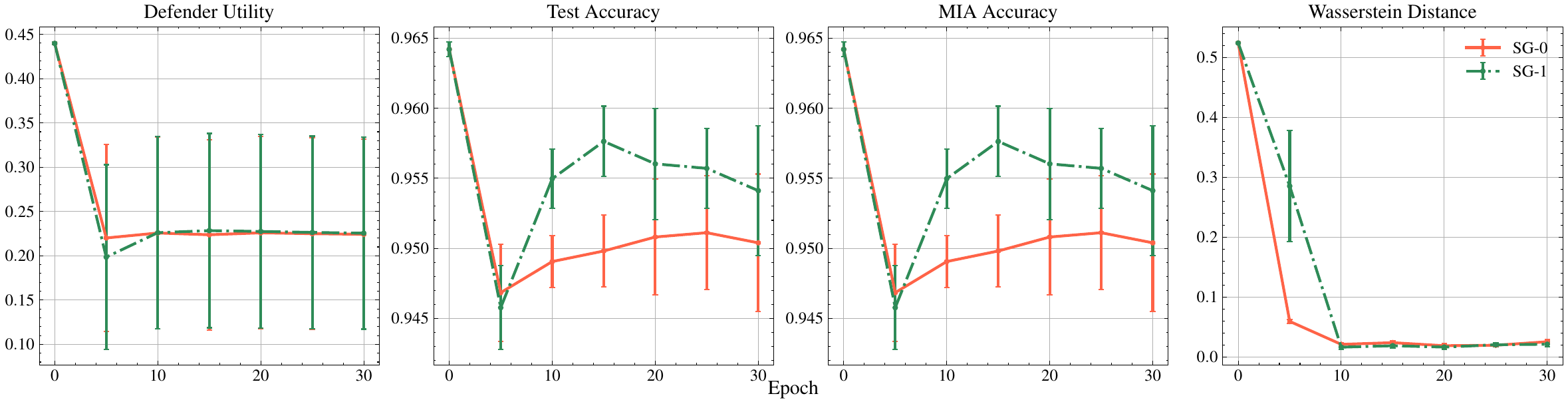}
    \label{fig:ablation_svhn}
\end{figure}

The comparison of SG on SVHN for random forgetting with and without attacker is illustrated in Figure \ref{fig:ablation_svhn}.

\subsubsection{Loss Distributions}
\begin{sidewaysfigure}
\includegraphics[width=\textwidth,]{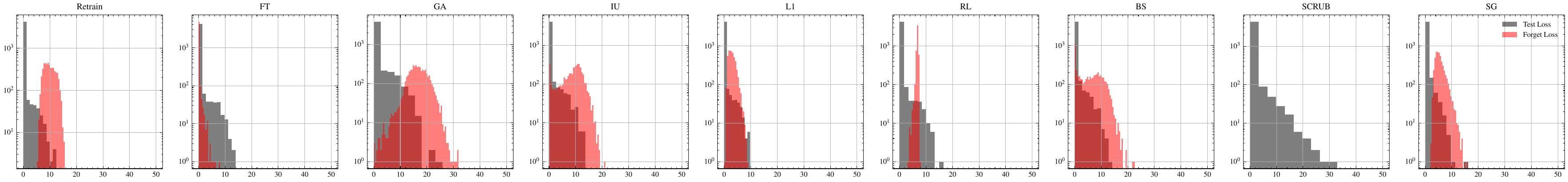}
\includegraphics[width=\textwidth,]{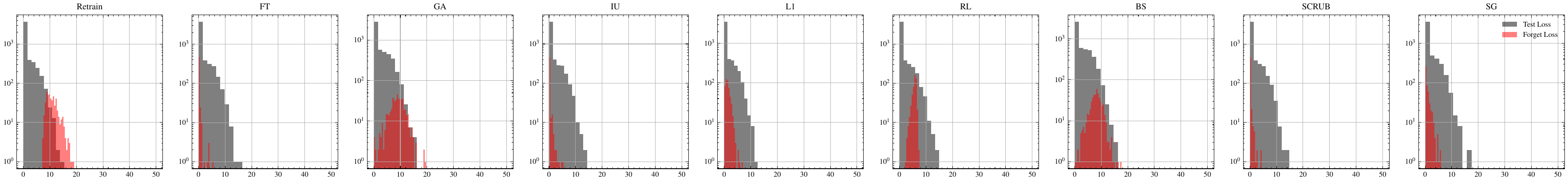}
\includegraphics[width=\textwidth,]{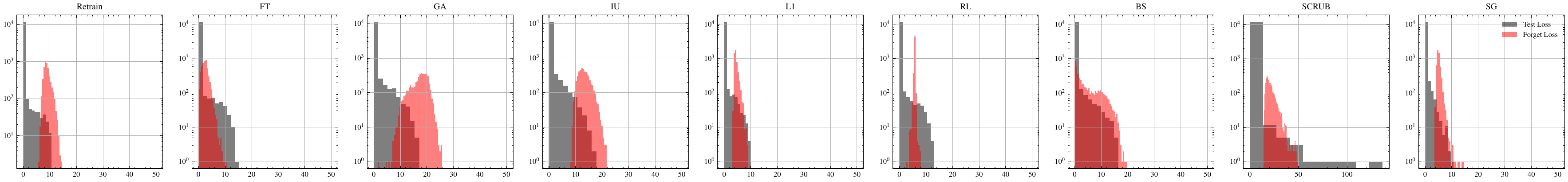}
\includegraphics[width=\textwidth,]{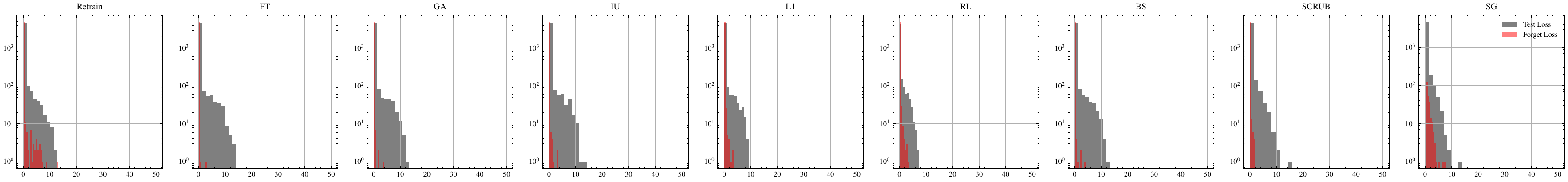}
\includegraphics[width=\textwidth,]{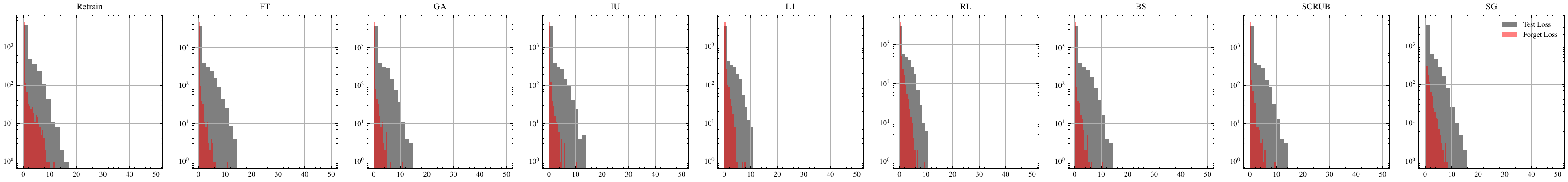}
\includegraphics[width=\textwidth,]{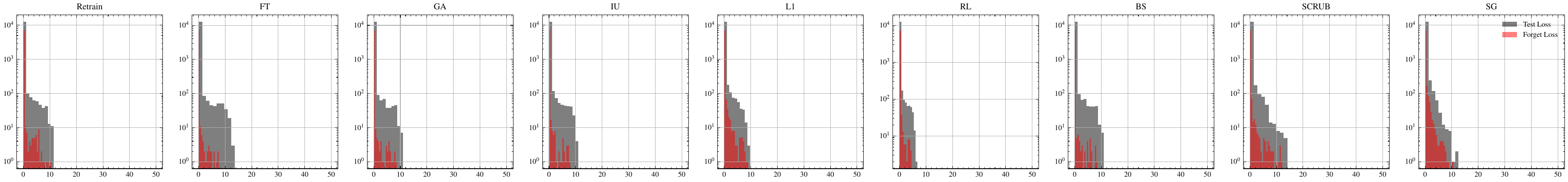}
\caption{The distributions of the cross-entropy losses for the forget and test instances from the unlearned models. The y-axis is in log scale for better visualization. From the first to the last figure, they are random forgetting on CIFAR-10, CIFAR-100, SVHN and class-wise forgetting on CIFAR-10, CIFAR-100, SVHN.}
\label{fig:loss_dist}
\end{sidewaysfigure}

A visualization of the cross-entropy losses of the forget and test instances is in Figure~\ref{fig:loss_dist}.

\subsubsection{Baseline Methods}
\label{sec:baseline_methods}

\begin{itemize}[itemsep=-0.2em,leftmargin=*,label={},topsep=0em, partopsep=0em]
    \itemTitle{Retrain} The first baseline is retraining, where the unlearned model is obtained by training on the retain set from scratch. 
    We aim to develop unlearning algorithms so that the metrics they produce are as closely aligned with those of the retraining as possible.
    \itemTitle{Fine-Tuning (FT)} As the second baseline, FT continues to train the original model  on the retain set for a few epochs. This a standard baseline used in various prior research~\citep{graves21amne,Warnecke23mu}.
    \itemTitle{Gradient Ascent (GA)} This baseline takes the original model as the starting point and runs a few epochs of gradient ascent  on the forget set \fdata. The intuition is to disrupt the model's generalizability on \fdata~\citep{graves21amne}. Another name of GA is NegGrad~\citep{kurmanji2024towards}.
    \itemTitle{Influence Unlearning (IU)} This baseline uses Influence Function to estimate the updates required for a model's weights as a result of removing the forget set from the training data~\citep{izzo21deletion,koh18inf}.
    \itemTitle{$\ell_1$-sparse} This baseline integrates an $\ell_1$ norm-based sparse penalty into machine unlearning loss \cite{jia23mu}. 
    \itemTitle{Random Label (RL)} This baseline trains the original model on the retain set and the forgetting set $D_f$ whose labels are random to make the model unlearn $D_f$ while keep the model capability as much as possible.
    \itemTitle{Boundary Expansion} This baseline proposes a neighbor searching method to identify the nearest but incorrect class labels to guide the way of boundary shifting. 
    \itemTitle{Boundary Shrink} This baseline artificially assigns forgetting samples to an extra shadow class of the original model \cite{chen2023boundary}. 
    \itemTitle{SCRUB} This baseline achieve MU by using a teacher model and student model \cite{kurmanji2024towards}. 
\end{itemize}

\subsubsection{Experiment Details}
\label{sec:exp_details}
\begin{figure}[ht] 
    \centering 
    \includegraphics[width=0.5\textwidth]{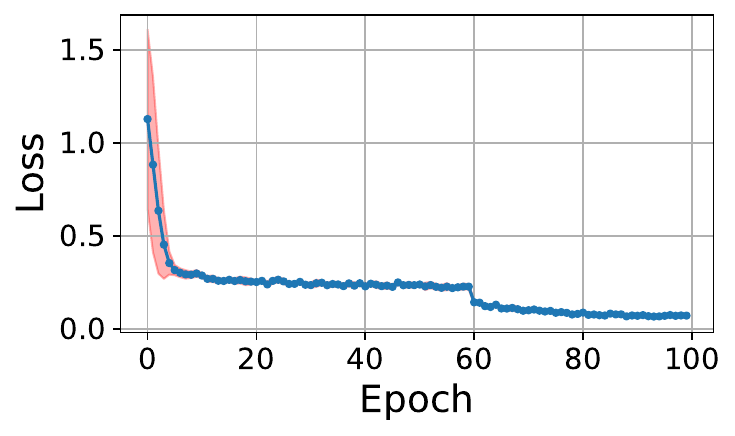}
    \includegraphics[width=0.5\textwidth]{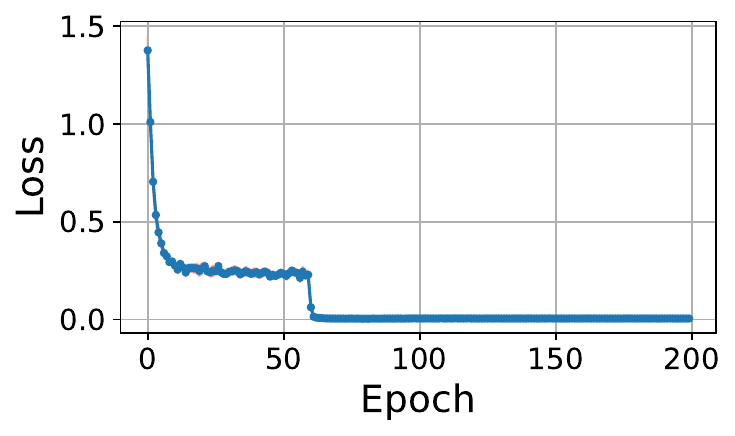}
    \includegraphics[width=0.5\textwidth]{figures/retrain_loss_svhn.pdf}
    \caption{
        The training loss for the retrain baseline. 
        For CIFAR10 and CIFAR100, the learning rate is multiplied by 0.1 when epoch is at 60, 120, 160; for SVHN, the same multiplication is done at epoch 60, 120. 
        \textbf{Top to bottom}: CIFAR10, CIFAR100, SVHN.
    }
    \label{fig:retrain_loss}
\end{figure}

The hyperparameters used for SG and the baselines are in Table~\ref{tab:hyperparameters}.
The losses for the retraining baseline across the epochs are displayed in Figure~\ref{fig:retrain_loss}.
We run all the experiments using PyTorch 1.12 on NVIDIA A5000 GPUs and AMD EPYC 7513 32-Core Processor.

\begin{table}[]
    \centering
    \caption{The hyper-parameter for the baseline method and SG used in this paper.}
    \scalebox{1}{
    \begin{tabular}{c|cccccccccc}
    \toprule
    Parameters & Retrain & FT & GA & IU & $\ell_1$-sparse & RL & BE & BS & SCRUB & SG \\
    \midrule
    Learning rate & 1e-2 & 5e-2 & 1e-3 & $\times$ & 1e-2 & 1e-2 & 1e-5 & 1e-5 & 5e-4 & 1e-2\\
    Num. of epoch & 160  & 30 & 5  & $\times$ & 10 & 10 & 10 & 10 & 10 & 30\\
    $\gamma$ & $\times$ & $\times$ & $\times$ & $\times$ & 5e-4 & $\times$ & $\times$ & $\times$ & $\times$ & $\times$ \\
    $\alpha$ & $\times$ & $\times$ & $\times$ & 10 & $\times$ & $\times$ & $\times$ & $\times$ & $\times$ & $\times$ \\
    $T$ &  $\times$ & $\times$ & $\times$ & $\times$ & $\times$ & $\times$ & $\times$ & $\times$ & 4 & $\times$ \\
    Decay epochs &  $\times$ & $\times$ & $\times$ & $\times$ & $\times$ & $\times$ & $\times$ & $\times$ & [3, 5, 9] & $\times$\\
    $\beta$ & $\times$ & $\times$ & $\times$ & $\times$ & $\times$ & $\times$ & $\times$ & $\times$ & 0.1 & $\times$ \\
    Attacker $\alpha$ & $\times$ & $\times$ & $\times$ & $\times$ & $\times$ & $\times$ & $\times$ & $\times$ & $\times$ & 1.0 \\
    \bottomrule
    \end{tabular}
    }
    \label{tab:hyperparameters}
\end{table}

\subsection{Class-wise Forgetting}
\label{app:class_wise}
The results of SVHN dataset on class-wise forgetting are given in Table \ref{tab:appendix_classwise}.

\begin{table}[ht]
    \centering
    \caption{Experimental results (Mean$_{\text{std}}$) on SVHN for classwise forgetting. The highlighted metrics are the closest to those of retraining, which is considered as the best performance compared with the other baselines.}
    \scalebox{0.65}{
    \begin{tabular}{c|cccc|cccccc}
    \toprule
     SVHN & ${\Acc}_{r}$  & ${\Acc}_{te}$ & ${\Acc}_{f}$ & $\Abs{{\Acc}_{f} - {\Acc}_{te}}$ & MIA acc. & MIA AUC & MIA F1 & KS Stat. & W. Dist. & RTE (min., $\downarrow$) \\
     \midrule
     Retrain & 0.9963$_{0.0001}$ & 0.9639$_{0.0007}$ & 0.0000$_{0.0000}$ & 0.9639 & 0.9950$_{0.0005}$ & 0.9986$_{0.0004}$ & 0.9951$_{0.0005}$ & 0.9909$_{0.0003}$ & 8.6924$_{0.0750}$ & 20.46 \\
     \midrule
     FT & \textbf{0.9978$_{0.0002}$} & \textbf{0.9622$_{0.0020}$} & 0.0995$_{0.0179}$ & 0.8627  & \textbf{0.9945$_{0.0005}$} & \textbf{0.9985$_{0.0006}$} & \textbf{0.9946$_{0.0007}$} & 0.9533$_{0.0038}$ & 2.3220$_{0.0216}$ & 3.17 \\
     GA & 0.9444$_{0.0055}$ & 0.9144$_{0.0047}$ & \textbf{0.0000$_{0.0000}$} & 0.9144 & 0.9969$_{0.0004}$ & 0.9998$_{0.0000}$ & 0.9970$_{0.0002}$ & 0.9849$_{0.0012}$ & 16.4834$_{0.2720}$ & 0.98 \\
     IU & 0.8044$_{0.1177}$ & 0.8061$_{0.0978}$ & \textbf{0.0000$_{0.0000}$} & 0.8061 & 0.9998$_{0.0003}$ & 1.0000$_{0.0000}$ & 0.9998$_{0.0003}$ & \textbf{0.9936$_{0.0056}$} & 15.0697$_{1.7117}$ & 0.41 \\
     $\ell_1$-sparse & 0.9799$_{0.0004}$ & 0.9580$_{0.0017}$ & \textbf{0.0000$_{0.0000}$} & 0.9580 & 0.9921$_{0.0013}$ & 0.9966$_{0.0003}$ & 0.9921$_{0.0012}$ & 0.9818$_{0.0030}$ & 4.5139$_{0.2512}$ & 3.73 \\
     RL & 0.9959$_{0.0001}$ & 0.9612$_{0.0013}$ & \textbf{0.0000$_{0.0000}$} & 0.9612 & 0.9912$_{0.0013}$ & 0.9971$_{0.0016}$ & 0.9913$_{0.0012}$ & 0.9813$_{0.0016}$ & \textbf{5.5978$_{0.0357}$} & 2.60 \\
     BE & 0.9880$_{0.0008}$ & 0.9546$_{0.0012}$ & 0.2812$_{0.0061}$ & 0.6734 & 0.9976$_{0.0005}$ & 0.9995$_{0.0002}$ & 0.9976$_{0.0006}$ & 0.9106$_{0.0050}$ & 4.3816$_{0.0659}$ & 0.46 \\
     BS & 0.9864$_{0.0010}$ & 0.9537$_{0.0010}$ & 0.3109$_{0.0052}$ & 0.6428 & 0.9975$_{0.0003}$ & 0.9995$_{0.0002}$ & 0.9976$_{0.0003}$ & 0.9072$_{0.0031}$ & 4.4290$_{0.1169}$ & 0.82 \\
     SCRUB & 0.9916$_{0.0007}$ & 0.9616$_{0.0014}$ & \textbf{0.0000$_{0.0000}$} & \textbf{0.9616} & 0.9999$_{0.0001}$ & 1.0000$_{0.0001}$ & 0.9999$_{0.0001}$ & 0.9989$_{0.0008}$ & 24.4590$_{2.2852}$ & 3.91 \\
     \midrule
     SG & 0.9716$_{0.0007}$ & 0.9601$_{0.0014}$ & \textbf{0.0000$_{0.0000}$} & 0.9601 & 0.9928$_{0.0001}$ & 0.9954$_{0.0001}$ & 0.9929$_{0.0001}$ & 0.9907$_{0.0008}$ & 5.0148$_{2.2852}$ & 5.92 \\
     \bottomrule
    \end{tabular}
    }
    \label{tab:appendix_classwise}
\end{table}








\subsection{An example of the condition in \eqref{eq:kkt}}\label{app:implicit_diff}
In this section, we provide a concrete example of the KKT conditions for linear support vector machines (SVM).
As described in Section~\ref{algo:sg_algo}, the KKT conditions are key to relating the attacker's model parameters, denoted as \attackerF, with the auditing set $\MIAdata_{\unlearnF}$, which allows us to derive the gradient $\partial \attackerF / \partial \MIAdata_{\unlearnF}$.
The conditions \KKT can be similarly derived for any model where the learning problem is convex.
To simplify the notations, we use  $\SET{(x_i, y_i)}_{i=1}^{\MIAnum}$ to represent $\MIAdata_{\unlearnF}$.
A standard formulation of the linear SVM is as follows
            \begin{equation}\label{eq:svm_opt}
                \begin{aligned}
                    & \min_{\attackerF, b} & & \frac{1}{2}\norm{\attackerF }^2 \\
                    & s.t.             & & y_i \cdot (\attackerF^\top x_i + b) \ge 1, \forall i,
                \end{aligned}
            \end{equation}
where $b$ is the bias term.
The standard form is typically formulated as a minimization problem, so the attacker is to maximize $\Lobj = - \frac{1}{2}\norm{\attackerF }^2$.
Eq.~\eqref{eq:svm_opt} is a convex program, and the optimal solution (i.e., $\attackerF^\ast$ and $b^\ast$) is characterized by the KKT conditions. 
The Lagrangian of the above is as follows where $\alpha_i \ge 0$ are the Lagrantian multipliers:
            \begin{equation}
                L(\attackerF, b, \alpha_i) = \frac{1}{2}\norm{\attackerF}^2 - \sum_{i=1}^{\MIAnum}{\alpha_i \left( y_i \cdot (\attackerF^\top x_i + b) - 1\right)}.
            \end{equation}
Following sandard procedures~\citep{boyd2004convex}, the KKT conditions are as folllows
            \begin{equation}\label{eq:svm_kkt}
            f(\MIAdata_{\unlearnF}, \attackerF) = \begin{cases}
                    \begin{aligned}
                    & \attackerF - \sum_{i=1}^{\MIAnum}{\alpha_i y_i x_i} = 0 \\
                    & -\sum_{i=1}^{\MIAnum}{\alpha_i y_i} = 0 \\
                    & y_i \cdot (\attackerF^\top x_i + b) \ge 1 \\
                    & \alpha_i \ge 0, \forall i \\
                    & \alpha_i ( y_i (\attackerF^\top x_i + b) - 1) = 0, \forall i
                    \end{aligned}
                \end{cases},
            \end{equation}
which implicitly define a function between \attackerF and the data $\MIAdata_{\unlearnF} = \SET{(x_i, y_i)}_{i=1}^{\MIAnum}$.
In practice, we describe the optimization problem \eqref{eq:svm_opt} using \texttt{cvxpy}~\citep{diamond2016cvxpy}.
Then, we employ an off-the-shelf package called \texttt{cvxpylayers}~\citep{Agrawal2019diffcvx} to automatically derive the KKT conditions and compute the gradient  $\partial \attackerF / \partial \MIAdata_{\unlearnF}$.
